\documentclass[runningheads]{llncs}

\usepackage{eccv}

\usepackage{multirow}

\usepackage{eccvabbrv}

\usepackage{tabularx}
\usepackage{graphicx}
\usepackage{booktabs}

\usepackage[accsupp]{axessibility}  

\usepackage[pagebackref,breaklinks,colorlinks,citecolor=eccvblue]{hyperref}
\usepackage{hyperref}

\usepackage{orcidlink}

\begin{document}

\title{Lifelong Person Re-Identification with Backward-Compatibility} 

\titlerunning{Lifelong Person Re-Identification with Backward-Compatibility}

\author{Minyoung Oh \and
	Jae-Young Sim
	}

\authorrunning{M. Oh and J.Y. Sim}

\institute{Ulsan National Institute of Science and Technology, Republic of Korea}

\maketitle

\begin{abstract}
Lifelong person re-identification (LReID) assumes a practical scenario where the model is sequentially trained on continuously incoming datasets while alleviating the catastrophic forgetting in the old datasets.	However, not only the training datasets but also the gallery images are incrementally accumulated, that requires a huge amount of computational complexity and storage space to extract the features at the inference phase. In this paper, we address the above mentioned problem by incorporating the backward-compatibility to LReID for the first time. We train the model using the continuously incoming datasets while maintaining the model's compatibility toward the previously trained old models without re-computing the features of the old gallery images.	To this end, we devise the cross-model compatibility loss based on the contrastive learning with respect to the replay features across all the old datasets. Moreover, we also develop the knowledge consolidation method based on the part classification to learn the shared representation across different datasets for the backward-compatibility. We suggest a more practical methodology for performance evaluation as well where all the gallery and query images are considered together. Experimental results demonstrate that the proposed method achieves a significantly higher performance of the backward-compatibility compared with the existing methods. It is a promising tool for more practical scenarios of LReID.
	
\end{abstract}

\section{Introduction}\label{sec:intro}
Person re-identification (ReID) aims to retrieve specific individuals (query) within the scene images collected from different cameras (gallery). It has drawn much attention in various practical applications such as security systems. The conventional ReID methods have been mainly developed in the supervised learning manner~\cite{reid1,reid2,reid3,PCB}. 
However, in real-world applications, ReID system environments are subject to dynamic changes unlike the conventional scenarios. For example, diverse datasets are continuously collected at different time instances, and the ReID models should learn the knowledge from these incrementally changing datasets to maintain the performance across diverse domains.

One of the trivial methods to address this problem is to train the model by using the whole datasets together, however it requires a huge computational complexity as well as increases the memory space for training datasets. Recently, Lifelong ReID (LReID) has been introduced as a solution to address these challenges~\cite{AKA, GwFReID, PTKP, KRKC}, which sequentially trains the model on continuously incoming datasets of diverse domains while reducing the computational complexity and storage demands.
At the same time, it attempts to alleviate the catastrophic forgetting, where the model prefers the knowledge learned from the recent dataset while forgetting the old knowledge learned from the previously given datasets.

However, in practice, the challenges lie not only on the training datasets but also on the accumulated gallery images for inference.
Whenever the model is updated with a new training dataset in LReID scenario, the feature space is changed from that with the old datasets.
This causes the model to re-compute the features from all the accumulated gallery images again to ensure that the features are compatible with the features of the query images extracted by the new model. We call this process as {\it backfilling}. 
Since the gallery images are also continuously collected, this backfilling process takes a substantial amount of time especially for long-term deployment of ReID systems.
However, the existing LReID methods do not consider this issue.

On the other hand, the concept of backward-compatible training (BCT) has been introduced to address the time-consuming task of backfilling~\cite{BCT, NCCL, AdvBCT}.
In the BCT scenario, the model is trained to ensure the backward-compatibility toward the previously trained old models such that the feature spaces are compatible between the new and old models.
Specifically, the BCT methods train the model by forcing the features computed by the new model to be close to the features obtained by the old model. Hence, based on BCT, the gallery features extracted by the old model can be re-used for re-identification with the new query features obtained by the new model, alleviating the burden of backfilling.
However, the existing BCT methods jointly train all the accumulated datasets to update the model and do not consider the domain shift between the training datasets.

\begin{figure}[t!]
	\centering
	\subfloat{\includegraphics[width=1\linewidth]{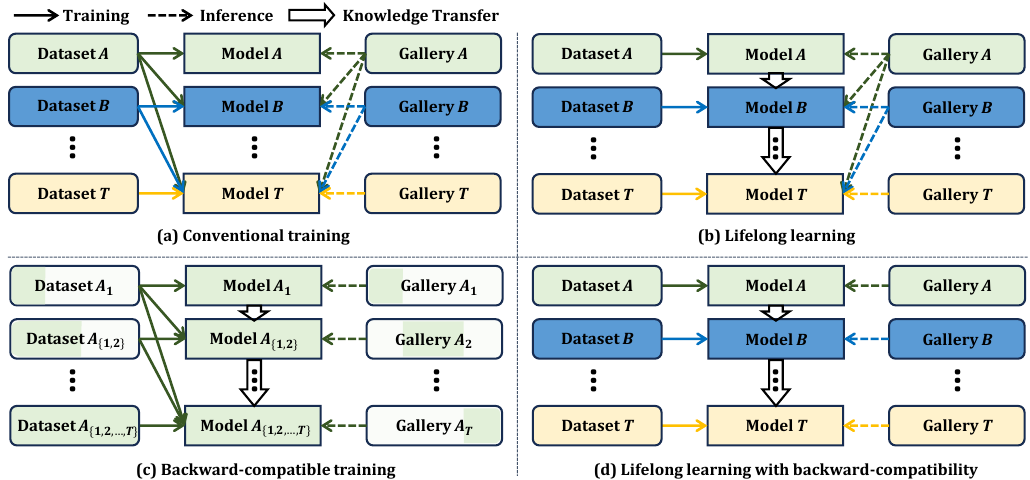}}
	\caption{The concept of the proposed lifelong person re-identification with backward-compatibility. (a) Conventional training. (b) Lifelong learning. (c) Backward-compatible training. (d) Lifelong learning with the backward-compatibility.}
	\label{fig:concept}
\end{figure} 

Based on the investigation of the complementary relationship between LReID and BCT, we propose a novel LReID framework combined with the backward-compatibility. 
\cref{fig:concept} shows the proposed framework compared to the other existing frameworks. (a) The conventional methodology trains the current model with all the accumulated datasets which is used to test all the accumulated gallery datasets. (b) The lifelong learning methodology sequentially trains the model in order by using each training dataset, but tests all the gallery datasets. (c) The existing backward-compatible training concept considers a certain dataset and trains the model by using a partial subset of the dataset. However, it does not re-compute the features of the old gallery data. (d) The proposed framework combines the concepts of (b) and (c) together, which sequentially trains the model with incoming new datasets in order but does not re-compute the features of the old gallery data.

As we sequentially train the model on incrementally incoming datasets in a lifelong learning manner, we ensure that the model's feature space remains compatible with that of the old models, thereby maintaining the backward-compatibility.
To this end, we propose the backward-compatible learning scheme. Based on the replay technique, we store a set of the replay images selected from the current dataset and their features extracted by the current model, referred to the replay features.
When training the new model, the features of the replay images extracted by the new model are guided by the cross-model compatibility loss that maximizes the similarity to their corresponding replay features while minimizing the similarity to the other ones. By doing so, the new model well represents the feature space of the old models.
On the other hand, ReID datasets usually do not share common identities that makes the models hard to transfer the knowledge learned from one dataset to the other.
We use the part classifier to encourage the model to convey the shared knowledge by utilizing the consistent local characteristics of the person images across different datasets.
Moreover, we consolidate the cross-attention by selectively leveraging the feature channels helpful to both the new and old tasks, leading the model to learn the shared representation between both models.
Experimental results show that the proposed method yields much better performance of LReID compared with the existing methods while achieving the backward-compatibility.

Our main contributions are summarized as follows:
\begin{itemize}
    \item To the best of our knowledge, we first introduce the framework of lifelong person re-identification considering backward-compatibility, which is a more practical scenario free from the backfilling process.
   
    \item We propose the cross-model compatibility learning scheme based on the contrastive learning utilizing the replay images and the features. We also propose the part-assisted knowledge consolidation method that selectively exploits the feature channels beneficial for both the new and old models.
    
    \item We devise a more practical but challenging evaluation methodology that considers all the query and gallery images from the entire datasets simultaneously. Experimental results on the benchmark datasets show that the superiority of the proposed method in terms of the backward-compatibility compared with the existing methods.

\end{itemize}

\section{Related Works}{
\subsection{Person Re-Identification}
ReID has made a great progress with impressive performance in a supervised manner.
However, there are three limitations of the conventional ReID methods. 
(1) Huge burden of labeled dataset acquisition:
Acquiring the labeled datasets for ReID is time-consuming and labor-intensive, due to the small inter-class variations and the large intra-class variations.
To alleviate the data acquisition burden, the unsupervised domain-adaptive methods~\cite{uda1,uda2}, fully-unsupervised methods~\cite{us1,us2}, and domain generalizable methods~\cite{dg1,dg2} have been studied.
(2) Limited characteristics of the existing datasets: 
Most of the conventional datasets for ReID include the day-time person images wearing a same cloth.
To handle more diverse and practical situations, the visible-infrared~\cite{vireid1,vireid2,vireid3}, clothes-changing~\cite{ccreid1,ccreid2,ccreid3}, and bird-view~\cite{birdview} setups have been introduced.
(3) Knowledge forgetting in old domains: 
In real-world scenarios, diverse datasets are continuously incoming, and thus the model training with all the datasets requires significant amount of the computational complexity as well as the memory space.
LReID methods~\cite{AKA, GwFReID, PTKP, KRKC} have been investigated to avoid the catastrophic forgetting problem while mitigating the burden of computation and memory.

\subsection{Lifelong Person Re-Identification}
LReID assumes the training datasets are continuously collected where each dataset represents a unique domain.
AKA~\cite{AKA} utilizes the learnable knowledge graph to effectively accumulate and transfer the knowledge. 
GwFReID~\cite{GwFReID} mitigates the catastrophic forgetting by ensuring the coherence of knowledge between the old and new models.
PTKP~\cite{PTKP} regards the LReID as a domain-adaptation problem where the old datasets are regarded as the source domain and the new dataset as the target domain.
KRKC~\cite{KRKC} keeps the old model trainable to refresh the old knowledge, which helps the model to consolidate the knowledge between the old and new models.
CLUDA-ReID~\cite{CLUDA} proposes an unsupervised domain-adaptation scenario in LReID.
However, the existing methods consider the continuously incoming training datasets only, overlooking that the gallery images are also accumulated increasing the computational burden as well as requiring memory consumption.

\subsection{Backward-Compatible Training}
The purpose of BCT is to train the new model to be compatible with the old models in terms of the feature extraction. It aims to avoid the time-consuming feature re-computation for all the gallery images whenever the model is updated with new datasets.
BCT~\cite{BCT} employs the old classifier and forces the features obtained by the new model to be compatible with the old classifier.
LCE~\cite{LCE} introduces the transformation module which maps the features into the other model's feature spaces.
NCCL~\cite{NCCL} enhances the backward-compatibility by using the contrastive learning with the features and classification logits of the new and old models.
AdvBCT~\cite{AdvBCT} introduces the adversarial learning which makes the features obtained by different models indistinguishable.
These methods are conducted under the assumption that all the training datasets are accessible during the training of the new model, which requires huge computational complexity and  storage space.
Recently, CVS~\cite{CVS} maintains the backward consistency of the models following the lifelong learning.
However, it does not consider dynamically changing datasets which usually cause huge distribution-shift. Moreover, the performance is evaluated on the classes seen in the training only, which lacks the applicability for practical scenarios.
}

\section{Methodology}

\begin{figure}[t]
    \centering
    \subfloat{\includegraphics[width=1\linewidth]{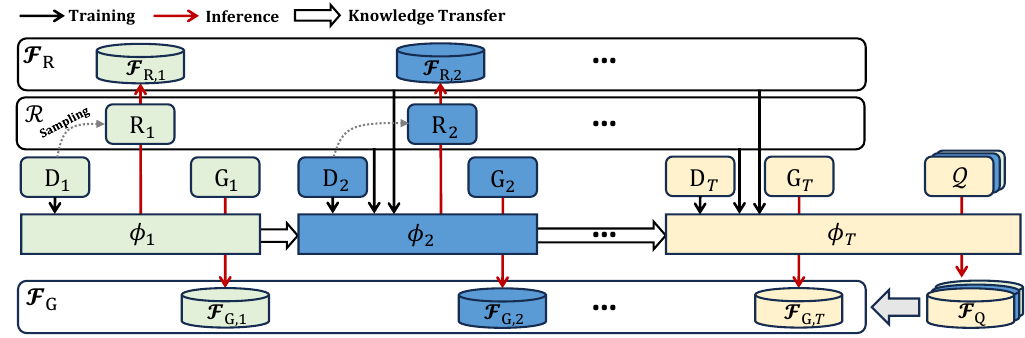}}
    \caption{Illustration of the training and inference procedures in the proposed framework.}
    \label{fig:pipeline}
\end{figure}

\subsection{Problem Formulation}
\cref{fig:pipeline} illustrates the training and inference procedures of the proposed LReID framework with backward-compatibility.
Following the conventional LReID formulation, let us assume that $T$ different datasets are given in order. We sequentially train the model $\phi$ (feature extractor) using the training datasets denoted as $\mathcal{D}=\{\mathrm{D}_{t} \}^{T}_{t=1}$.
We test the model on the gallery datasets $\mathcal{G}=\{\mathrm{G}_t\}^T_{t=1}$ with respect to the query sets $\mathcal{Q}=\{\mathrm{Q}_t\}^T_{t=1}$.
At the $t$-th time instance, we train the model $\phi_t$ by using $\mathrm{D}_t$ and sample the replay data $\mathrm{R_t}$ from $\mathrm{D_t}$. Note that the total replay datasets are given as $\mathcal{R}=\{\mathrm{R_t}\}^{T-1}_{t=1}$ since we do not need to sample the replay data at the last dataset. We extract the features of the replay data such that $\boldsymbol{\mathcal{F}}_{\mathrm{R},t}=\{\phi_t(\boldsymbol{r}_{t,i})\}^{|\mathrm{R_t}|}_{i=1}$ where $\boldsymbol{r}_{t,i}$ represents the $i$-th replay image in $\mathrm{R_t}$.
In addition, we also compute the features of the gallery data for inference purpose, which are stored into  $\boldsymbol{\mathcal{F}}_{\mathrm{G},t} = \{\phi_t(\boldsymbol{g}_{t,i})\}^{|\mathrm{G_t}|}_{i=1}$ where $\boldsymbol{g}_{t,i}$ represents the $i$-th gallery image in $\mathrm{G_t}$. 
Note that, in the LReID scenario, the features of the gallery images are obtained as $\boldsymbol{\mathcal{F}}_{\mathrm{G},t} = \{\phi_T(\boldsymbol{g}_{t,i})\}^{|\mathrm{G_t}|}_{i=1}$.
After the training of the last ($T$-th) model, we compute the features $\boldsymbol{\mathcal{F}}_\mathrm{Q}$ of the query images from $\mathcal{Q}$ by using $\phi_T$. Then we evaluate the performance of LReID on the set of the pre-computed gallery features $\boldsymbol{\mathcal{F}}_\mathrm{G} = \left\{\boldsymbol{\mathcal{F}}_{\mathrm{G},t}\right\}^{\mathrm{T}}_{t=1}$. Since we do not need to re-compute the features of gallery datasets, it is a more practical setup without requiring the time-consuming backfilling process.

\subsection{Baseline Network Training}
We construct the baseline composed of the feature extractor $\phi$ and the classifier $\psi$.
We adopt cross-entropy loss for identity discriminative capability, given by
\begin{equation}
    \mathcal{L}_\mathrm{ce} = - \frac{1}{|B|}\sum_{i\in B}\sum_{j\in Z}{{y}_{i,j}\operatorname{log}\left({p}_{i,j}\right)},
    \label{eq:ce}
\end{equation}
where $B$ is a single mini-batch, $Z$ denotes the set of the identity labels, and ${y}_{i,j}$ is the one-hot label for the $i$-th image $\boldsymbol{x}_i$ with respect to the $j$-th label.
${p}_{i,j}$ represents the probability that the $i$-th image has the $j$-th identity label, which is the output of the softmax operation on $\psi(\phi(\boldsymbol{x}_i))$.
We also use the hard triplet loss~\cite{triplet} with the Euclidean distance given by
\begin{equation}
    \mathcal{L}_\mathrm{tri}=\frac{1}{|B|}\sum_{i\in B}\operatorname{max}\left(||\boldsymbol{f}_i - \boldsymbol{f}^{i+}||_2 - ||\boldsymbol{f}_i - \boldsymbol{f}^{i-}||_2 + m, 0\right),
    \label{eq:triplet}
\end{equation}
where $\boldsymbol{f}_i$ is the output feature vector of $\phi(\boldsymbol{x}_i)$ which serves as an anchor, $\boldsymbol{f}^{i+} \text{ and } \boldsymbol{f}^{i-}$ denote the hardest positive and negative feature vectors with respect to $\boldsymbol{f}_i$ in a mini-batch, and $m$ is a hyper-parameter specifying the margin. 
For the training of the baseline network, we use the loss $\mathcal{L}_\mathrm{base}$ as
\begin{equation}
    \mathcal{L}_\mathrm{base} = \lambda_\mathrm{ce}\mathcal{L}_\mathrm{ce} + \lambda_\mathrm{tri}\mathcal{L}_\mathrm{tri},
    \label{eq:base}
\end{equation}
where $\lambda_\mathrm{ce}$ and $\lambda_\mathrm{tri}$ are hyper-parameters.

\subsection{Cross-Model Compatibility Loss}
Whenever the model is sequentially trained on new datasets, the resulting feature spaces are changed to exhibit unique distributions according to the training domains.
In such cases, $\phi_t$ becomes incompatible with the previous models $\{\phi_1, \phi_2, ..., \phi_{t-1}\}$, and we usually  backfill $\boldsymbol{\mathcal{F}}_\mathrm{G}$ with the features of $\mathcal{G}$ extracted by $\phi_t$ to compute the feature similarity between $\boldsymbol{\mathcal{F}}_\mathrm{G}$ and $\boldsymbol{\mathcal{F}}_\mathrm{Q}$.

To ensure $\phi_t$ to be compatible with the previously trained models, we design the cross-model compatibility loss inspired by the concept of the supervised contrastive loss~\cite{supcon}, given by 
\begin{equation}
	\!\!\!\!\!\!\!\!\mathcal{L}_\mathrm{cmcl} \!\! =\! -\frac{1}{|B^r|}\!\! \sum_{i \in B^r} \!\! \operatorname{log}\! \left(\! \frac{ \sum^{t-1}_{l=1} \sum_{j=1}^{|\boldsymbol{\mathcal{F}}^{i+}_{\mathrm{R},l}|} \exp\left\{\frac{\phi_t(\boldsymbol{r}_i)\cdot \boldsymbol{f}^{i+}_{l,j}}{\tau} \right\} } 
	{\sum^{t-1}_{l=1}\!\! \sum_{j=1}^{\boldsymbol{|\mathcal{F}}_{\mathrm{R},l}|}\! \exp\!\left\{\!\!\frac{\phi_t(\boldsymbol{r}_i)\cdot \boldsymbol{f}_{l,j}}{\tau}\!\!\right\}
	\!+\! \sum_{j\in B}\! \exp\!\left\{\!\!\frac{\phi_t(\boldsymbol{r}_i)\cdot \phi_t(\boldsymbol{x}_j)}{\tau}\!\!\right\} } \!\right)\!\!, \!\!\!\!\!\!	
	\label{eq:cmcl}
\end{equation}
where $B^r$ is a mini-batch sampled from $\mathcal{R}$, $\boldsymbol{\mathcal{F}}^{i+}_{\mathrm{R},l}$ is the set of the positive replay features to $\boldsymbol{r}_i$ of $\boldsymbol{\mathcal{F}}_{\mathrm{R},l}$, $\boldsymbol{f}^{i+}_{l,j}$ is the $j$-th replay feature in $\boldsymbol{\mathcal{F}}^{i+}_{\mathrm{R},l}$, $\boldsymbol{f}_{l,j}$ is the $j$-th replay feature in $\boldsymbol{\mathcal{F}}_{\mathrm{R},l}$, and $\tau$ is a temperature parameter.
Note that the supervised contrastive loss~\cite{supcon} takes the positive and negative samples within a mini-batch only without considering the model's backward-compatibility. On the contrary, we exploit the replay features across all the old datasets up to the $(t-1)$th time instance via $\left\{\boldsymbol{\mathcal{F}}_{\mathrm{R},l}\right\}^{t-1}_{l=1}$ based on the lifelong learning framework to preserves the backward-compatibility more reliably. 
Moreover, based on the assumption that different datasets do not include the same identity simultaneously, we minimize the similarity between the feature of the old replay image $\boldsymbol{r}_i$ and the feature of the new training image $\boldsymbol{x}_i$ extracted by the new model $\phi_t$. Consequently, $\phi_t$ gets better representation by exploiting more negative features from the perspective of the contrastive learning, as discussed in \cite{simclr}.

There has been an attempt~\cite{NCCL} to achieve the backward-compatibility by using the supervised contrastive loss, which uses $\mathcal{R}$ and the old model $\phi_{t-1}$ instead of $\boldsymbol{\mathcal{F}}_\mathrm{R}$ where $\phi_{t-1}(\boldsymbol{r}_i)$ is utilized to guide the new model to represent the feature space obtained by $\phi_{t-1}$.
However, in the proposed lifelong learning formulation, the training datasets are dynamically changing across long duration of time instances, and thus it is hard for $\phi_{t-1}$ to keep the knowledge of the feature space at very old time instances suffering from the backward-compatibility of $\phi_t$ towards the old models.
We note that the features of $\boldsymbol{\mathcal{F}}_{\mathrm{R}}$ are extracted by the old models maintains the representation of the previously trained models, and thus employ them to guide the new model $\phi_t$ to ensure the backward-compatibility.

\subsection{Part-Assisted Knowledge Consolidation}\label{sec:3-4}{
Unlike the formulation of the existing BCT methods~\cite{BCT,NCCL,LCE,AdvBCT}, we assume the lifelong learning framework where the model sequentially trains different datasets.
The absence of the common supervision for model training makes it hard to learn the shared representation across the different datasets and the model's backward-compatibility accordingly.
To learn the shared representation, we adopt the part features that exhibit the common characteristics of the human parts.
Note that several ReID methods~\cite{PCB, part2} also utilized the part features to learn the part-aware knowledge. However, we aim to learn the shared part features for the purpose of backward-compatibility.

We add a part pooling layer that is parallel to the global pooling layer in $\phi_t$. 
Similar to PCB~\cite{PCB}, the part pooling layer horizontally divide the feature map into $N$ partitions, $\{\boldsymbol{f}^{\mathrm{p}[n]}\}^{N}_{n=1}$.
Each part feature is assigned the part label and fed into the part classifier $\varphi_t$, which is trained by the part classification loss
\begin{equation}
    \mathcal{L}_\mathrm{pcl} = - \frac{1}{|B|} \frac{1}{N} \sum_{i\in B} \sum_{n=1}^{N} \sum_{m=1}^{N} {y}^{\mathrm{p}[n]}_{i,m}\operatorname{log}\left({p}^{\mathrm{p}[n]}_{i,m}\right),
\end{equation}
where ${y}^{\mathrm{p}[n]}_{i,m}$ is the one-hot label corresponding to the $n$-th part feature $\boldsymbol{f}^{\mathrm{p}[n]}_i$ with respect to the $m$-th part feature, and ${p}^{\mathrm{p}[n]}_{i,m}$ is the probability that $\boldsymbol{f}^{\mathrm{p}[n]}_i$ is extracted from the $m$-th part, which is the output of the softmax operation of $\varphi_t(\boldsymbol{f}^{\mathrm{p}[n]}_i)$. 
Note that while the new identities from different datasets change the model's feature space drastically degrading the backward-compatibility, a common objective across the different datasets helps the model to yield consistent feature spaces by regularizing the model not to be overly biased to the task-specific features.

We additionally utilize the part classifier $\varphi_{t-1}$ of the old model and freeze the parameters of $\varphi_{t-1}$ to preserve the old feature space. 
We simultaneously minimize $\mathcal{L}_\mathrm{pcl}$ using $\varphi_{t-1}(\boldsymbol{f}^{\mathrm{p}[n]}_i)$.
Note that $\varphi_{t-1}$ is jointly trained with $\phi_{t-1}$, and thus their feature spaces are highly correlated in general. 
This implies that minimizing $\mathcal{L}_\mathrm{pcl}$ using $\varphi_{t-1}(\boldsymbol{f}^{\mathrm{p}[n]}_i)$ enhances the backward-compatibility toward $\phi_{t-1}$.
However, since the fixed parameters of $\varphi_{t-1}$ are optimized to $\mathrm{D}_{t-1}$, the feature space of $\varphi_{t-1}$ is different from that of $\varphi_{t}$.
It means that minimizing $\mathcal{L}_\mathrm{pcl}$ jointly for both $\varphi_{t}(\boldsymbol{f}^{\mathrm{p}[n]}_i)$ and $\varphi_{t-1}(\boldsymbol{f}^{\mathrm{p}[n]}_i)$ could lead to the conflict when optimizing the parameters of $\phi_t$.

\begin{figure}[t]
    \centering
    \subfloat{\includegraphics[width=1\linewidth]{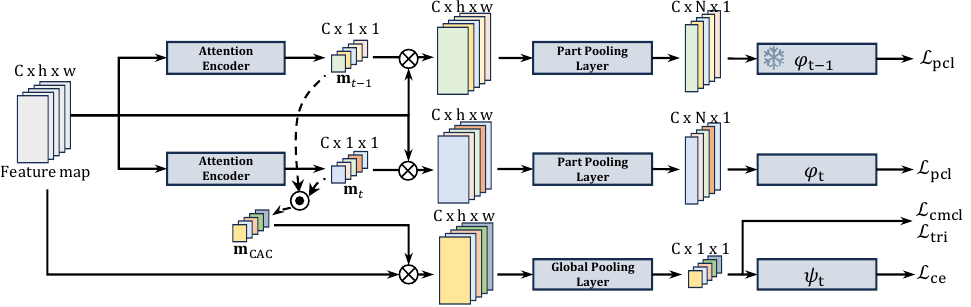}}
    \caption{Part-assisted knowledge consolidation. 
    }
    \label{fig:PACOL}
\end{figure}

Therefore, as shown in~ \cref{fig:PACOL}, we employ two independent channel-wise attention encoders coupled with $\varphi_{t-1}$ and $\varphi_{t}$, respectively. 
Two encoders yield the attention masks $\boldsymbol{\mathrm{m}}_{t-1}$ and $\boldsymbol{\mathrm{m}}_t$ along the channels that highlight the significant channels independently.
We estimate the cross-attention consolidation $\boldsymbol{\mathrm{m}}_\mathrm{CAC}$.
\begin{equation}
    \boldsymbol{\mathrm{m}}_\mathrm{CAC} = \boldsymbol{\mathrm{m}}_{t-1} \odot \boldsymbol{\mathrm{m}}_{t},
    \label{eq:consolidation}
\end{equation}
where $\odot$ indicates the element-wise multiplication. 
$\boldsymbol{\mathrm{m}}_\mathrm{CAC}$ can be used for learning the shared representation for both tasks at $t-1$ and $t$ while mitigating the conflict between both tasks by selectively emphasizing the channels that are crucial for both part classifiers.
Then we apply $\boldsymbol{\mathrm{m}}_\mathrm{CAC}$ to the feature map before the global pooling layer for the identity discriminative learning.

\section{Experimental Results}

\subsection{Experimental Setup}
\subsubsection{Datasets.}
To verify the effectiveness of the proposed method, we used four benchmark datasets with unique domains: Market1501\cite{Market}, DukeMTMC\cite{Duke}, CUHK-SYSU\cite{CUHK}, and MSMT17\cite{MSMT}, as widely used in the existing literatures of LReID~\cite{GwFReID, PTKP, ConRFL}. 
Since CUHK-SYSU\cite{CUHK} is introduced for person search, which takes the scene images as input, we pre-process it for the purpose of ReID using the ground-truth annotations following~\cite{GwFReID, PTKP, KRKC}. The specifications of the used datasets are shown in~\cref{tab:dataset}.

\begin{table}[t]
    \centering
    \caption{The statistics of datasets.}
    \begin{tabular}{>{\centering\arraybackslash}p{0.15\textwidth}|> {\centering}p{0.09\textwidth} |> {\centering}p{0.16\textwidth} |> {\centering}p{0.16\textwidth} |>{\centering}p{0.16\textwidth}|>{\centering\arraybackslash}p{0.16\textwidth}}
    \hline
                \multicolumn{2}{c|}{}                   & Market1501 & DukeMTMC & CUHK-SYSU & MSMT17\\ \hline
                \multirow{2}{*}{Training}     & Images    & 12,936& 16,522& 15,088& 32,621\\ \cline{2-6}
                                            & ID        & 751& 702& 5,532& 1,041\\ \hline
                \multirow{2}{*}{Query}     & Images    & 3,368& 2,228 & 2,900& 11,659\\ \cline{2-6}
                                            & ID        & 750& 702& 2,900& 3,060\\\hline
                \multirow{2}{*}{Gallery}   & Images    & 15,913& 17,661& 5,447& 82,161\\ \cline{2-6}
                                            & ID        & 751& 1,110& 2,900& 3,060\\\hline
    \end{tabular}
    \label{tab:dataset}
\end{table} 

\subsubsection{Evaluation Protocol.}\label{sec:eval}
As illustrated in \cref{fig:pipeline}, we compute all the query features $\{\boldsymbol{\mathcal{F}}_\mathrm{Q,t}\}^T_{t=1}$ by using $\phi_T$.
On the other hand, each set of gallery features in $\{\boldsymbol{\mathcal{F}}_\mathrm{G,t}\}^T_{t=1}$ is pre-computed by using the corresponding model of each task $\{\phi_t\}^T_{t=1}$.
Whereas the existing LReID methods require the backfilling process with respect to $\boldsymbol{\mathcal{F}}_\mathrm{G}$ by using $\phi_T$, the proposed evaluation protocol represents more practical scenarios without the backfilling process.
To evaluate the performance, we adopt the mean average precision (mAP) score and the Rank-1 (R-1) accuracy.

\subsubsection{Implementation Details.}
We adopted the ImageNet pre-trained ResNet50~\cite{resnet} with some modifications as the feature extractor. 
The global average pooling layer is replaced with the generalized mean (GeM) pooling layer~\cite{GeM} following~\cite{PTKP}. 
The two part pooling layers are added in parallel with the GeM pooling layer. 
Moreover, the two independent attention encoders, introduced in \cref{sec:3-4}, are adopted before each part pooling layer. 
All the networks $\phi_t$, $\varphi_t$ and $\psi_t$ are jointly trained except for $\varphi_{t-1}$ at the training phase, however, only $\phi_t$ is employed for inference phase.
Our total loss function to train $\phi_t$ is formulated as
\begin{equation}
	\mathcal{L}_\mathrm{total} = \mathcal{L}_\mathrm{base} + \lambda_\mathrm{cmcl}\mathcal{L}_\mathrm{cmcl} + \lambda_\mathrm{pcl}\mathcal{L}_\mathrm{pcl},
	\label{eq:total}
\end{equation}
where $\lambda_\mathrm{cmcl}$ and $\lambda_\mathrm{pcl}$ are hyper-parameters.
Hyper-parameters are empirically set as follows: $m=0.3$ in \cref{eq:triplet}, $\lambda_\mathrm{ce}=1$ and $\lambda_\mathrm{tri}=1$ in \cref{eq:base}, $\tau=0.5$ in \cref{eq:cmcl}, $N=5$ for part classification, and $\lambda_\mathrm{cmcl}=0.1$ and $\lambda_\mathrm{pcl}=1$ in \cref{eq:total}.
We followed \cite{PTKP} for other training details such as the replay memory update strategy, learning rate, etc.

\subsection{Performance Comparison} \vspace{-1mm}
To compare the performance of LReID with backward-compatibility, we re-implement several existing methods: LwF\cite{LwF}, BCT\cite{BCT}, CVS\cite{CVS}, PTKP\cite{PTKP}, and KRKC\cite{KRKC}. 
\cref{tab:comparison} shows the results of the proposed method compared with that of the existing method following the evaluation protocol explained in \cref{sec:eval}.
Joint-train means that we trained the model on all the datasets together, and Fine-tuning means that the model is independently trained on each new dataset in order. Both methods are trained with $\mathcal{L}_\mathrm{base}$ only.

\begin{table}[t]
    \centering
    \caption{Quanitative performance comparison. We re-implemented the existing methods to evaluate the performance of the backward-compatibility.}
    \begin{tabular}{c|cc|cc|cc|cc|cc}
         \hline
         \multicolumn{11}{c}{ Train Order: Market1501 $\rightarrow$ DukeMTMC $\rightarrow$ CUHK-SYSU $\rightarrow$ MSMT17 }  \\ \hline
         \multirow{2}{*}{Method} & \multicolumn{2}{c|}{Market1501} & \multicolumn{2}{c|}{DukeMTMC} & \multicolumn{2}{c|}{CUHK-SYSU} & \multicolumn{2}{c|}{MSMT17} & \multicolumn{2}{c}{Average}\\ \cline{2-11}
          & mAP & R-1 & mAP & R-1 & mAP & R-1 & mAP & R-1 & mAP & R-1\\ \hline\hline
         Joint-train & 81.1 & 93.4 & 71.1 & 86.0 & 91.8 & 93.2 & 41.1 & 69.8 & 71.3 & 85.6\\ 
         Fine-tuning & 59.3 & 80.3 & 54.2 & 74.7 & 82.3 & 84.7 & 38.6 & 67.6 & 58.6 & 76.8\\ \hline
         LwF\cite{LwF} & 62.9 & 82.0 & 56.8 & 74.8 & 78.8 & 81.2 & 13.5 & 33.7 & 53.0 & 67.9\\ 
         BCT\cite{BCT} & 75.1 & 89.7 & 60.2 & 78.0 & 84.6 & 86.3 & 29.8 & 56.6 & 62.4 & 77.7\\ 
         CVS\cite{CVS} & 70.7 & 87.0 & 56.3 & 72.0 & 81.5 & 83.8 & 27.2 & 53.0 & 58.9 & 74.0\\ 
         KRKC\cite{KRKC} & 59.5 & 80.3 & 49.3 & 69.7 & 80.7 & 82.4 & \textbf{36.7} & \textbf{65.3} & 56.6 & 74.4\\ 
         PTKP\cite{PTKP} & 73.4 & 86.4 & 62.7 & 77.8 & 83.8 & 85.3 & 36.3 & 62.0 & 64.1 & 77.9\\ \hline
         Proposed & \textbf{80.4} & \textbf{92.3} & \textbf{64.6} & \textbf{80.7} & \textbf{87.3} & \textbf{88.9} & 34.8 & 63.3 & \textbf{66.8} & \textbf{81.3} \\ \hline
    \end{tabular}
    \label{tab:comparison}
    \vspace{0mm}
\end{table} 

We observe that the performance of all the methods on MSMT17, the last dataset, degrades compared to that of Fine-tuning due to the stability-plasticity dilemma~\cite{dilemma} that the bias on the old model's knowledge disrupts the adaptation capability of the new knowledge.
LwF suffers from the biasing toward the old knowledge and yields poor performance on MSMT17. 
On the one hand, PTKP achieves noticeable performance although it does not explicitly consider the backward-compatibility in training. 
Since PTKP casts the LReID problem into the domain adaption problem regarding the old dataset as source dataset, the model maps the features from the new dataset into the feature space of the old model that is trained with the source dataset.
Thus, it could enhance backward-compatibility of the new model.
On the other hand, KRKC also addresses the LReID problem but it trains the old model as well unlike the other methods that freeze the parameters of the old model while preserving its feature space. 
Hence it exhibits poor backward-compatibility while adapting the new knowledge with the best performance on MSMT17 among the other methods.
As the old model is trained, the feature space of old model is gradually forgotten, and the old model becomes no longer suitable for guiding the new model to ensure the backward-compatibility in KRKC.
It is noteworthy that the existing LReID methods, e.g., PTKP,  can enhance the backward-compatibility of the model, however 
it does not always guarantee the backward-compatibility as in the case of KRKC.

\begin{table}[t]
    \centering
    \caption{Performance of the proposed method in different training orders.}
    \begin{tabular}{c|cc|cc|cc|cc|cc}
         \hline
         \multicolumn{11}{c}{ Train Order: Market1501 $\rightarrow$ CUHK-SYSU $\rightarrow$ DukeMTMC $\rightarrow$ MSMT17}  \\ \hline
         \multirow{2}{*}{Method} & \multicolumn{2}{c|}{Market1501} & \multicolumn{2}{c|}{CUHK-SYSU} & \multicolumn{2}{c|}{DukeMTMC} & \multicolumn{2}{c|}{MSMT17} & \multicolumn{2}{c}{Average}\\ \cline{2-11}
          & mAP & R-1 & mAP & R-1 & mAP & R-1 & mAP & R-1 & mAP & R-1\\ \hline\hline
         Fine-tuning & 56.6 & 78.0 & 72.6 & 74.5 & 57.4 & 76.0 & \textbf{38.1} & \textbf{67.5} & 56.2 & 74.0\\ 
         Proposed & \textbf{80.0} & \textbf{92.0} & \textbf{85.3} & \textbf{87.1} & \textbf{64.8} & \textbf{79.7} & 33.9 & 61.7 & \textbf{66.0} & \textbf{80.1}\\ \hline
         \multicolumn{11}{c}{}  \\ \hline
         \multicolumn{11}{c}{ Train Order: DukeMTMC $\rightarrow$ Market1501 $\rightarrow$ CUHK-SYSU $\rightarrow$ MSMT17}  \\ \hline
         \multirow{2}{*}{Method} & \multicolumn{2}{c|}{DukeMTMC} & \multicolumn{2}{c|}{Market1501} & \multicolumn{2}{c|}{CUHK-SYSU} & \multicolumn{2}{c|}{MSMT17} & \multicolumn{2}{c}{Average}\\ \cline{2-11}
          & mAP & R-1 & mAP & R-1 & mAP & R-1 & mAP & R-1 & mAP & R-1\\ \hline\hline
         Fine-tuning & 56.5 & 75.4 & 58.1 & 80.0 & 80.8 & 82.8 & \textbf{38.6} & \textbf{67.8 }& 58.5 & 76.5\\ 
         Proposed & \textbf{71.5} & \textbf{85.6} & \textbf{71.1} & \textbf{88.9} & \textbf{85.8} & \textbf{87.8} & 32.9 & 61.6 & \textbf{65.3} & \textbf{81.0}\\ \hline
    \end{tabular}
    \label{tab:orders2}
    \vspace{4mm}
\end{table} 

To verify the robustness of the proposed method to arbitrary training orders, we additionally conducted the experiments in different training orders. \cref{tab:orders2} shows the results where we see that the proposed method improves the performance by a large margin from that of the existing methods, regardless of the training orders.

Furthermore, we devise a novel approach to evaluate the performance which is a more practical yet challenging scenario that takes all the pre-computed gallery features of $\boldsymbol{\mathcal{F}}_\mathrm{G}$ into account with respect to all the query features of $\boldsymbol{\mathcal{F}}_{\mathrm{Q}}$ computed by $\phi_T$.
As shown in \cref{tab:gallery_all}, the proposed method outperforms the other methods in terms of both mAP score and R-1 accuracy.
\cref{fig:qual} also compares the qualitative results of Top-10 matching. We see that the proposed method successfully finds the query person whereas the other methods often return the incorrect matching results.

\subsection{Ablation Study}
To verify the effectiveness of each component in the proposed method, we conducted the ablation study as shown in \cref{tab:ablation}. 
It shows that each component of the proposed method contributes to the performance gain in an additive manner.

\begin{table}[t]
	\centering
	\caption{Performance when taking all the gallery sets $\boldsymbol{\mathcal{F}}_\mathrm{G}$ into account for evaluation with respect to all of query features $\boldsymbol{\mathcal{F}}_{\mathrm{Q}}$.}
	\begin{tabular}{cc|cc|cc|cc|cc|cc|cc}\hline
		\multicolumn{14}{c}{ Train Order: Market1501 $\rightarrow$ DukeMTMC $\rightarrow$ CUHK-SYSU $\rightarrow$ MSMT17 }  \\ \hline
		\multicolumn{2}{c|}{Fine-tuning} & \multicolumn{2}{c|}{LwF\cite{LwF}} & \multicolumn{2}{c|}{BCT\cite{BCT}} & \multicolumn{2}{c|}{CVS\cite{CVS}} & \multicolumn{2}{c|}{KRKC\cite{KRKC}} & \multicolumn{2}{c|}{PTKP\cite{PTKP}} & \multicolumn{2}{c}{Proposed} \\ \hline
		mAP & R-1 & mAP & R-1 & mAP & R-1 & mAP & R-1 & mAP & R-1 & mAP & R-1 & mAP & R-1 \\ \hline\hline
		40.6 & 63.6 & 22.7 & 37.3 & 43.0 & 65.3 & 35.2 & 54.3 & 36.4 & 55.9 & 47.0 & 66.5 & \textbf{48.9} & \textbf{71.5} \\ \hline
	\end{tabular} \vspace{0mm}
	\label{tab:gallery_all}
\end{table} 

\begin{figure}[t]
	\centering
	\subfloat{\includegraphics[width=1\linewidth]{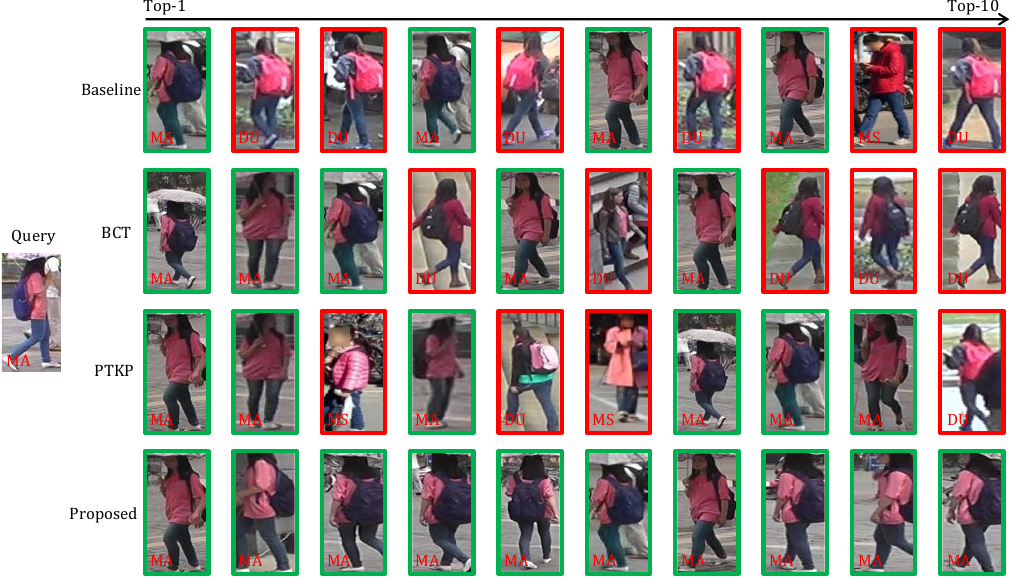}}
	\caption{Visualization of the Top-10 matching results of the proposed method compared with that of the existing methods, where all gallery features of $\boldsymbol{\mathcal{F}}_\mathrm{G}$ and all query features of $\boldsymbol{\mathcal{F}}_\mathrm{Q}$ are considered simultaneously. The true and false matching results are respectively depicted in the green and red colors, respectively.
		MA: Market1501, DU: DukeMTMC, MS: MSMT17.
	}
	\label{fig:qual}
	\vspace{-4mm}
\end{figure} 

\begin{table}[t]
	\centering
	\caption{Ablation study of each component in the proposed method. CAC: cross-attention consolidation.}
	\begin{tabular}{ccc|cc|cc|cc|cc|cc}\hline
		\multicolumn{13}{c}{ Train Order: Market1501 $\rightarrow$ DukeMTMC $\rightarrow$ CUHK-SYSU $\rightarrow$ MSMT17}  \\ \hline
		\multicolumn{3}{c|}{Method}& \multicolumn{2}{c|}{Market1501} & \multicolumn{2}{c|}{DukeMTMC} & \multicolumn{2}{c|}{CUHK-SYSU} & \multicolumn{2}{c|}{MSMT17} & \multicolumn{2}{c}{Average}\\ \hline
		$\mathcal{L}_\mathrm{cmcl}$ &$\mathcal{L}_\mathrm{pcl}$ &$\mathrm{CAC}$& mAP & R-1 & mAP & R-1 & mAP & R-1 & mAP & R-1 & mAP & R-1\\ \hline\hline
		& &                                      & 59.3 & 80.3 & 54.2 & 74.7 & 82.3 & 84.7 & 38.6 & 67.6 & 58.6 & 76.8\\ \hline
		\checkmark&  &                           & 77.4 & 90.9 & 62.6 & 78.6 & 86.0 & 87.9 & 33.7 & 61.7 & 64.9 & 79.8\\ \hline
		&  \checkmark&                           & 60.3 & 81.2 & 56.2 & 75.4 & 82.4 & 84.6 & \textbf{39.3} & 68.7 & 59.6 & 77.5\\ \hline
		&  \checkmark& \checkmark     & 65.5 & 84.0 & 57.3 & 76.5 & 84.3 & 86.3 & 39.1 & \textbf{68.8} & 61.6 & 78.9\\ \hline
		\checkmark&  \checkmark & \checkmark    & \textbf{80.4} & \textbf{92.3} & \textbf{64.6} & \textbf{80.7} & \textbf{87.3} & \textbf{88.9} & 34.8 & 63.3 & \textbf{66.8} & \textbf{81.3}\\ \hline
	\end{tabular}
	\label{tab:ablation}
	\vspace{4mm}
\end{table} 

\begin{figure}[t]
    \centering
    \subfloat{\includegraphics[width=0.8\linewidth]{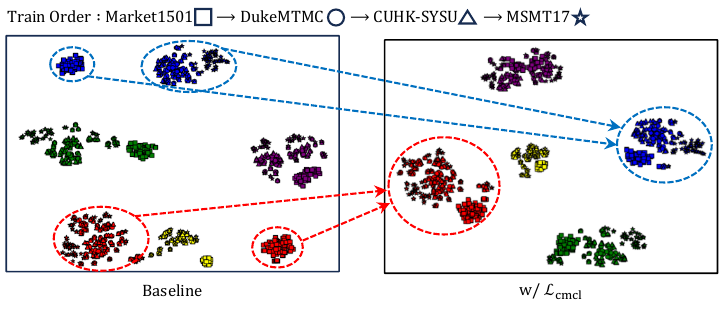}}
    \caption{Visualization of the feature distributions on Market1501. Different colors indicate different identities and different shapes represent the features extracted from different models, respectively.
    }
    \label{fig:tsne}
    \vspace{4mm}
\end{figure} 

\subsubsection{Effectiveness of Cross-Model Compatibility Loss.}
\cref{tab:ablation} shows that $\mathcal{L}_\mathrm{cmcl}$  significantly improves the performance of the model's backward-compatibility. 
\cref{fig:tsne} also visualizes the feature distribution on Market1501 by using t-SNE\cite{TSNE} to show how $\mathcal{L}_\mathrm{cmcl}$ affects the feature distribution. 
Each color represents each person identity, and different shapes indicate that the features are computed from different models, for example, the squares are features computed by $\phi_1$, and the circles are obtained by $\phi_2$. 
Some features are separated by the baseline although they have the same identities as shown in the dashed circles.
However, we observe that the features of a same identity get closer to each other comprising a compact cluster.

\begin{table}[t]
    \centering
    \caption{Performance variation according to the value of $\tau$ in $\mathcal{L}_\mathrm{cmcl}$.}
    \begin{tabular}{c|cc|cc|cc|cc|cc}
         \hline
         \multicolumn{11}{c}{ Train Order: Market1501 $\rightarrow$ DukeMTMC $\rightarrow$ CUHK-SYSU $\rightarrow$ MSMT17}  \\ \hline
         \multirow{2}{*}{$\mathcal{L}_\mathrm{cmcl}$} & \multicolumn{2}{c|}{Market1501} & \multicolumn{2}{c|}{DukeMTMC} & \multicolumn{2}{c|}{CUHK-SYSU} & \multicolumn{2}{c|}{MSMT17} & \multicolumn{2}{c}{Average}\\ \cline{2-11}
           & mAP & R-1 & mAP & R-1 & mAP & R-1 & mAP & R-1 & mAP & R-1\\ \hline\hline
         $\tau=0.1$ & \textbf{79.4} & \textbf{91.5} & 62.4 & \textbf{78.9} & \textbf{86.8} & \textbf{88.4} & 31.0 & 58.5 & \textbf{64.9} & 79.3\\ \hline
         $\tau=0.5$ & 77.4 & 90.9 & \textbf{62.6} & 78.6 & 86.0 & 87.9 & 33.7 & 61.7 & \textbf{64.9} & 79.8\\ \hline
         $\tau=1.0$ & 76.0 & 89.7 & 61.9 & \textbf{78.9} & 85.1 & 87.0 & 35.7 & 63.8 & 64.7 & 79.6\\ \hline
         $\tau=1.5$ & 75.0 & 89.6 & 60.8 & 78.2 & 85.1 & 87.0 & \textbf{36.5} & \textbf{65.3} & 64.4 & \textbf{80.0}\\ \hline
         \end{tabular}
    \label{tab:tau}
    \vspace{4mm}
\end{table} 

Moreover, \cref{tab:tau} shows additional experimental results to figure out how the performance changes according to the values of $\tau$.
We observe that when $\tau$ gets smaller, the performance on the oldest dataset improves while that on on the most recent dataset degrades.
Conversely, the model shows the better performance on the most recent dataset but degrades the performance on the oldest dataset with large values of $\tau$.
As discussed in \cite{temperature}, $\tau$ in the supervised contrastive loss, that our $\mathcal{L}_\mathrm{cmcl}$ is based on, determines the contribution of the hard samples to training.
We discuss it is because the features computed from the oldest dataset serve as the most hardest samples for model training.
While the model trains with continuously changing datasets, the feature space keeps changing according to the distinct characteristics of different datasets.
It disrupts the model's backward-compatibility towards the oldest model, making the oldest features in $\boldsymbol{\mathcal{F}}_\mathrm{R}$ used for $\mathcal{L}_\mathrm{cmcl}$ to be regarded as the hard samples.
Thus, small values of $\tau$ lead the older features to contribute more during the model training, and this yields better performance on the older datasets while accompanying slight performance degradation on relatively recent datasets.

\subsubsection{Effectiveness of Part-Assisted Knowledge Consolidation.}
\cref{tab:ablation} also shows the effect of the part-assisted knowledge consolidation method.
We see the performance gains when the part classification loss $\mathcal{L}_\mathrm{pcl}$ is applied to all the datasets, implying $\mathcal{L}_\mathrm{pcl}$ helps the model to learn the shared representation for backward-compatibility. 
Furthermore, the cross-attention consolidation (CAC) also contributes to improve the performance by a large margin.

\begin{table}[t]
    \centering
    \caption{Optimal design for CAC. All the experiments are conducted incorporating $\mathcal{L}_\mathrm{base}$ and $\mathcal{L}_\mathrm{pcl}$.}
    \begin{tabular}{ccc|cc|cc|cc|cc|cc}\hline
         \multicolumn{13}{c}{ Train Order: Market1501 $\rightarrow$ DukeMTMC $\rightarrow$ CUHK-SYSU $\rightarrow$ MSMT17}  \\ \hline
         \multicolumn{3}{c|}{Method}& \multicolumn{2}{c|}{Market1501} & \multicolumn{2}{c|}{DukeMTMC} & \multicolumn{2}{c|}{CUHK-SYSU} & \multicolumn{2}{c|}{MSMT17} & \multicolumn{2}{c}{Average}\\ \hline
         $\mathcal{L}_\mathrm{cmcl}$ & $\mathrm{CAC_A}$ & $\mathrm{CAC_M}$& mAP & R-1 & mAP & R-1 & mAP & R-1 & mAP & R-1 & mAP & R-1\\ \hline\hline
          &    \checkmark   &                   & 64.0 & 83.6 & 56.8 & 76.1 & 83.9 & 85.4 & \textbf{40.1} & \textbf{69.1} & 61.2 & 78.6\\ \hline
          \checkmark&\checkmark       &         & 78.1 & 91.3 & 63.4 & 80.4 & 86.3 & 88.0 & 35.7 & 64.2 & 65.9 & 81.0    \\ \hline
          && \checkmark&65.5 & 84.0 & 57.3 & 76.5 & 84.3 & 86.3 & 39.1 & 68.8 & 61.6 & 78.9\\ \hline
          \checkmark&  & \checkmark& \textbf{80.4} & \textbf{92.3} & \textbf{64.6} & \textbf{80.7} & \textbf{87.3} & \textbf{88.9} & 34.8 & 63.3 & \textbf{66.8} & \textbf{81.3}\\ \hline
    \end{tabular}
    \label{tab:cac-opt}
\end{table} 

On the other hand, the averaging of the two attention masks can be another option to obtain the consolidated masks for CAC such that 
\begin{equation}
\boldsymbol{\mathrm{m}}_\mathrm{CAC} = \frac{\boldsymbol{\mathrm{m}}_t + \boldsymbol{\mathrm{m}}_{t-1}} {2}.
\label{eq:cac-opt}
\end{equation}
To figure out the optimal design for CAC, we conducted the additional experiments as shown in \cref{tab:cac-opt}.
$\mathrm{CAC_M}$ and $\mathrm{CAC_A}$ indicate that $\boldsymbol{\mathrm{m}}_\mathrm{CAC}$ is obtained by \cref{eq:consolidation} and \cref{eq:cac-opt}, respectively.
We see that both options improve the performance with and without $\mathcal{L}_\mathrm{cmcl}$, however $\mathrm{CAC_M}$ contributes more on the performance gain except for the marginal difference on MSMT17 dataset. Since the values of $\boldsymbol{\mathrm{m}}_{t-1}$ and $\boldsymbol{\mathrm{m}}_{t}$ are bounded from 0 to 1, the values of $\mathrm{CAC_M}$ are assigned more conservatively compared to $\mathrm{CAC_A}$, addressing the optimization conflict problem more carefully.
Therefore, in this work, we took $\mathrm{CAC_M}$ for CAC implementation.

\section{Conclusion}
In this paper, we pointed out the limitation of the conventional LReID that considers the continuously incoming training datasets only and requires a significantly time-consuming backfilling of the features at the inference on the accumulated gallery images.
We proposed a novel LReID methodology that guarantees the backward-compatibility of the model to the old models without requiring the backfilling process.
We first designed the cross-model compatibility loss that utilizes the features of the replay images to guide the new model to represent the old feature space. 
We also proposed the knowledge consolidation scheme based on the part classification. It helps the model to learn the shared representation while mitigating the conflict between the old and new models.
Furthermore, we proposed a more practical yet challenging evaluation methodology for backward compatible LReID, that simultaneously takes all the gallery image and all the query images into account.
Experimental results showed that the proposed method enhances the backward-compatibility by a large margin compared with the existing methods.

%
%
\bibliographystyle{splncs04}
\bibliography{main}
\end{document}